\documentclass[10pt, journal] {IEEEtran}
\usepackage{floatrow}
\DeclareFloatFont{normalsize}{\normalsize}
\usepackage{amsmath}
\usepackage{graphicx}%
\usepackage{amsfonts}%
\usepackage{amssymb}
\usepackage{amsthm}  %
\usepackage{subfig}
\usepackage{tabularx}
\usepackage{multirow}
\usepackage{makecell}
\usepackage{adjustbox}
\usepackage{url}

\usepackage{caption}

\title{\huge CNN based Multistage Gated Average Fusion (MGAF) for Human Action Recognition Using Depth and Inertial Sensors}

\author{Zeeshan Ahmad, \textit{Student Member, IEEE}, Naimul Khan, \textit{Senior Member, IEEE}}

\begin{document}

	\maketitle

\begin{abstract}

Convolutional Neural Network (CNN) provides leverage to extract and fuse features from all layers of its architecture. However, extracting and fusing intermediate features from different layers of CNN structure is still uninvestigated for Human Action Recognition (HAR) using depth and inertial sensors. To get maximum benefit of accessing all the CNN's layers, in this paper, we propose novel Multistage Gated Average Fusion (MGAF) network which extracts and fuses features from all layers of CNN using our novel and computationally efficient Gated Average Fusion (GAF) network, a decisive integral element of MGAF. At the input of the proposed MGAF, we transform the depth and inertial sensor data into depth images called sequential front view images (SFI) and signal images (SI) respectively. These SFI are formed from the front view information generated by depth data. CNN is employed to extract feature maps from both input modalities.  GAF network fuses the extracted features effectively while preserving the dimensionality of fused feature as well. The proposed MGAF network has structural extensibility and can be unfolded to more than two modalities. 
Experiments on three publicly available multimodal HAR datasets demonstrate that the proposed MGAF outperforms the previous state-of-the-art fusion methods for depth-inertial HAR in terms of recognition accuracy while being computationally much more efficient. We increase the accuracy by an average of 1.5\% while reducing the computational cost by approximately 50\% over the previous state-of-art.

\end{abstract} 


\section{Introduction}
	
\IEEEPARstart{H}{UMAN} action recognition has progressed by leaps and bounds over the last two decades due to its many direct and indirect applications in numerous circles of our daily life such as media industry~\cite{samanta2014space}, robotics~\cite{chi2017gait}, sports~\cite{zhu2008event}, healthcare~\cite{corbishley2007breathing} and smart surveillance~\cite{dhiman2019robust}. These applications are indeed the true motivation behind the extensive research that has been carried out on HAR over the past years. \let\thefootnote\relax\footnote{© 2020 IEEE. Personal use of this material is permitted. Permission from IEEE must be obtained for all other uses, in any current or future media, including reprinting/republishing this material for advertising or promotional purposes, creating new collective works, for resale or redistribution to servers or lists, or reuse of any copyrighted component of this work in other works.}

Earlier, the research on HAR was on single modality sensor, either the vision sensor or inertial sensor. Vision sensors include RGB cameras and Kinect sensors while the inertial sensors for HAR are accelerometer and gyroscope. Using single modalities, several methods for HAR have been proposed over the years and these approaches set some promising directions for human action analysis~\cite{ramanathan2014human}~\cite{wang2019comparative}. However, each modality has limitations and deficiencies. For instance, conventional RGB cameras acquire 2D images and they are sensitive to lighting conditions, background clutter, partial occlusion and thus making action recognition task more challenging~\cite{wang2018rgb}.

The revolution in imagery technology emerges a new opportunity for HAR in the shape of  cost-effective depth sensors such as Kinect sensor which is capable of providing  3D action data, require less resources to operate and is less sensitive to lighting changes and illumination~\cite{aggarwal2014human}. Yet, there are limitations associated with depth images such as view point variation, sensitiveness to noise during image
acquisition and effect on the privacy of patients while using for healthcare applications. To address these limitations, researchers started to use wearable inertial sensors for HAR such as accelerometer and gyroscope~\cite{yang2009distributed}. Wearable sensors provide 3D action data in terms of multivariate time series. These sensors provide data at a high sampling rate and can work in gloomy and confounded conditions. Like vision based sensors, inertial sensors also have limitations such as sensor drift, inadequate onboard power and awkwardness of wearing them all the time~\cite{ehatisham2019robust}.

From above discussion it is clear that no single modality can encounter the challenging and unexpected situations that may arise in practice. The way out is to integrate the data from two modalities that can complement each other. Since depth images capture global features  while inertial
signals capture local  attributes, the fusion of these two modalities leads to robust and improved recognition~\cite{chen2017survey}. Furthermore, these two sensors are cost-effective, provide 3D action data and their multimodal fusion requires less computational complexities for HAR. We use depth images rather than the video camera images because the images obtained from video cameras are sensitive to lightening and illuminations, have limited viewing angle, needs aperture adjustments and require computationally complex processing algorithms than depth images~\cite{ehatisham2019robust}.

Earlier fusion methods for HAR using depth and inertial sensors were based on hand crafted methods where features are extracted using statistical methods like mean and variance in time domain~\cite{bao2004activity}
and the coefficients of Fast Fourier transform in frequency
domain~\cite{krause2003unsupervised}. The disadvantages with  classical methods are the requirement of domain knowledge about the data and the separation of feature extraction part from the classification part~\cite{plotz2011feature}. Furthermore, these hand crafted feature methods capture only a subset of the features, resulting in difficulty to generalize for unseen data. Outstanding performance of deep learning models, especially the convolutional neural network in the field of computer vision and image classification~\cite{krizhevsky2012imagenet}~\cite{ahmad2019humanactionrec}, trigger researchers to design deep learning based multimodal fusion frameworks for HAR using depth and inertial sensors~\cite{ahmad2019human}.

A major challenge in multimodal fusion is to decide when and where to integrate the modalities for optimal result~\cite{ramachandram2017deep}. Mulimodal fusion of sensors can be performed at feature level and decision.
The major disadvantage with decision level fusion is that it requires $N$ classifiers to train and test for $N$ modalities. Furthermore, with decision level fusion the correlated data necessary to improve recognition accuracy cannot be combined at earlier stages. In conradiction, the feature-level fusion ensures the collection and integration of correlated and concurrent information from the modalities necessary for the classifier to make a rigorous decision. Thus, feature level fusion assures the semantic information to flow from data to classifier~\cite{atrey2010multimodal}.

To address the aforementioned deficiencies and to exploit the fact that CNN allows to extract features from all the layers of its structure, in this paper, we present a novel multistage gated average fusion (MGAF) network capable of extracting and fusing features from all the layers of CNN using our proposed GAF network. GAF network effectively fuses the features from CNN layers using high boost kernel and gating mechanism. GAF network can be smoothly unrolled to more than two modalities without increasing the dimensionality of the fused features. 

The key novelty of the proposed MGAF resides in an intelligent approach of its gated average fusion (GAF) network that extracts, selects and fuses features in an efficient manner so that the accuracy of the classification task could be increased without increasing the computation cost as compared to the existing works that depend upon concatenation and the creation of the additional modalities.

The overview of the proposed MGAF network is shown in Fig.~\ref{fig:Proposed framework} and is explained in detail in section~\ref{sec:proposed method}.

 The novelty in the proposed framework can be summarized through the following three key contributions :
\begin{enumerate}	
	
	\item We propose a unique fusion network called multistage gated average fusion (MGAF) network which takes advantage of intermediate features of CNN and extracts discriminative, complementary and multiscale features from these layers. The extracted features are then fused using a novel gated average fusion (GAF) network which is an important componenet of MGAF. The proposed MGAF has architectural flexibility and can be stretched out to more than two modalities.

	\item Gated average fusion (GAF) network, which is the backbone of proposed MGAF, serves two purposes : (1) Fusing incoming feature maps from modalities by commissioning proper gated values to each pixel of feature maps. These values are calculated through gating mechanism, similar to the input gate function of long short-term memory (LSTM) network. (2) keeping the size of the fused feature map equal to the size of the feature map of single input modality by performing gated average fusion. Thus, the proposed GAF not only fuses the features but also performs dimensionality reduction.
	
	\item We experiment on three publicly available datasets with an efficient CNN consisting of only three convolutional layers. Achieving state of art outcomes using small CNN infers the excellence of proposed MGAF network over existing proposed fusion frameworks for depth-inertial HAR. We match or outperform all the state-of-the-art methods, including the frameworks recently proposed by us in~\cite{ahmad2019human}, while significantly reducing both training and inference computational cost.

\end{enumerate}	

The rest of the paper is organized as follows. Section II describes the related works on HAR using depth and inertial sensors individually and their conventional and deep learning based fusion. Section III provides detailed description of the proposed method. In section IV, we provide detailed experimental analysis, where the aforementioned contributions are analyzed in detail through large number of experiments and comparison with state-of-the-art models. Section V concludes the paper.

\section{Related Work}
\subsection{Depth data based approaches}
 Large number of methods for HAR using depth data have been proposed in the literature. A real time human action recognition system is developed in~\cite{chen2016realtim} using depth motion maps generated from depth sequences and $l_2$ regularized collaborative representation classifier. In~\cite{li2010action}, action graphs are generated from depth map sequences to model the dynamics of the actions and a bag of 3D points to characterize a set of salient postures that correspond to the nodes in the action graph. Authors in~\cite{yang2012recognizing} generate Histograms of Oriented Gradients (HOG) from depth motion maps to build DMM-HOG descriptors for human action recognition.  

 In~\cite{chen2017multi}, human action descriptor based on local spatio-temporal information from depth video sequences is proposed. This descriptor works in three stages and is capable of distinguishing similar actions performed with different speeds. A fast and accurate real time human pose recognition system is developed in~\cite{shotton2011real} that uses an intermediate body parts representation to map the difficult pose estimation problem into a simpler per-pixel classification problem. A comprehensive survey on human action recognition using depth data is provided in~\cite{liang2015survey} where the recent work and challenges related to depth modality are described in detail.

\subsection{Inertial and other sensor based approaches}
HAR using inertial sensors became popular due to their certain advantages over depth sensors such as insensitiveness against any lighting and illumination conditions. In~\cite{chen2015deep}, CNN is used to classify eight different human action from the data collected by tri-axial accelerometer. 
In~\cite{yan2019wiact}, extreme learning machine is used to classify ten actions using time series data obtained from WiFi sensor. In~\cite{jiang2015human}, wearable sensor data is converted into activity images and then CNN is employed for action classification. Experiments on three datasets show that proposed method is robust and highly accurate. In~\cite{tufek2019human}, limited data from accelerometer and gyroscope is used for HAR using CNN, LSTM and classical machine learning algorithms. Accuracy of action recognition was further increased by applying data augmentation and data balancing techniques. A survey article~\cite{wang2018deep} provides details about the performance of current deep learning models and future challenges on sensor based activity recognition. 

Other sensor modalities than inertial and depth for existing HAR are radar, sonar and video cameras. In~\cite{avci2010activity}, a survey on different applications of activity recognition such as healthcare, well being and sports using inertial sensors is presented. All the related challenges from data collection to classification and application directions are discussed in detail. 
In~\cite{sakuma2018wearable}, a wearable strain sensor technology is introduced for capturing and interpreting the dynamics of human fingertip. The
wearable device transmits raw deformation data to an off-finger device for interpretation. Simple
motions, gestures, finger-writing, grip strength, and activation time, as well as more complex idioms
consisting of multiple grips, are identified and quantified. Applications of proposed technology in different fields are also discussed.
In~\cite{blumrosen2016real}, the limitations of Kinect sensor is addressed by incorporating a set of features to create a unique “Kinect signature” for identification of different subjects during assessment of body kinematics of elderly people and patients with neurological disorders while performing daily life activities at home.
In~\cite{blumrosen2014noncontact}, wideband sonar is used to detect and classify human activities in indoor environment.
Wideband sonar is capable of tracking body parts precisely and enhanced correlation properties of sonar distinguish between human and non-human objects. The proposed technology can also be applied for monitoring patients at home and for detecting intruders.
In~\cite{li2019survey}, experimental based feasibility of applying  deep learning methods for  human activity recognition using radar data is surveyed. Current research challenges and future opportunities  in various fields such as human–computer interaction, smart surveillance and health assessment is discussed at length.
In~\cite{tan2018multi}, a novel RGB activity image-based DCNN classifier for the unobtrusive recognition of the multi-resident activities is proposed. 
The proposed method is experimentally proven to be helpful for increasing the recognition rate as compared to the previous studies.

\subsection{Fusion of depth and inertial data}
Some recent methods on HAR are based on fusing different modalities to get better recognition results. Human action recognition by fusing depth and inertial sensor data has gained significant attention due to their advantages over other modalities.
In~\cite{chen2015improving}, the accuracy of human action recognition is improved by fusing features extracted from depth and inertial sensor data and using collaborative representation classifier. Improved accuracy results were achieved due to complementary aspect of data from both modalities. A decision level fusion is performed between depth cameras and wearable sensors to increase the capabilities of robots to recognize human actions~\cite{manzi2018enhancing}. An efficient real-time human action recognition system
is developed in~\cite{chen2016real} using decision level fusion of depth and inertial sensor data. Depth and inertial data is effectively merged in~\cite{liu2014fusion} to train hidden Markov model for improving accuracy and robustness of hand gesture recognition.

 In~\cite{dawar2018real}, computationally efficient real-time detection and recognition approach is presented 
to identify actions in the smart TV application from continuous action streams using continuous integration of information obtained from depth and inertial sensor data. A novel, user friendly and safe method of bilateral gait segmentation is proposed in~\cite{hu2018novel} by multimodal fusion of features obtained from thigh mounted inertial sensor
 and depth sensor with the contralateral leg in its field
of view. The proposed method can be used to make lower limb assistive devices for patients with walking impairments. A comprehensive survey on fusion of depth and inertial sensors is provided in~\cite{chen2017survey} where the recent success of the fusion and future challenges and trends are disscussed in detail.

	\subsection{Deep Learning based Fusion}
	After the knock out performance of deep learning models on numerous machine learning and computer vision applications, several deep learning based fusion frameworks are presented.
	In~\cite{ahmad2018towards}, we proposed a CNN based fusion framework where depth and inertial sensor data is tranformed into images and then CNNs are employed to extract features from the transformed images. Finally concatenation fusion is performed between the features obtained from depth and signal images. These fused features served as input to train multiclass SVM classifier. Deep learning based decision level multimodal fusion framework is proposed in~\cite{dawar2019data}. CNN is used to extract features from depth images while Recurrent Neural Netwrok (RNN) is used to capture features from inertial sensor data. Data augmentation is also carried out to cop up with limited size data. In~\cite{bernal2018deep}, a supervised deep multimodal fusion framework for process monitoring and verification in the medical and healthcare fields is presented that depends on simultaneous processing of motion data acquired with wearable sensors and video data acquired with body-mounted camera. A CNN based sensor fusion system is developed in~\cite{dawar2018convolutional} to detect and monitor transition movements between body states as well as falls in healthcare applications. A score level sensor fusion is performed between the features extracted from depth and inertial sensor data by two CNNs.
	
	In~\cite{wei2019fusion}, simultaneously captured video and inertial data are converted into 3D video images and 2D inertial images respectively and are then fed as inputs into a 3D convolutional
	neural network and a 2D convolutional neural network, respectively, for recognizing actions. Then decision level and feature level fusion are performed to improve the classification accuracy. In~\cite{dawar2018action}, deep learning based fusion system
	based on fusing depth and inertial data is presented which
	is capable of detecting and recognizing actions of interest from
	continuous action streams. CNN is used to extract features
	from depth images, and a combination of CNN and LSTM network is utilized for inertial
	signals. First, segmentation is performed on each sensing
	modality and then actions of interest were detected. Finally, decision level fusion is performed for recognition. In~\cite{wei2020simultaneous}, video and inertial data, introduced in~\cite{wei2020c} is converted into 2D and 3D images which are used as input to 2D and 3D CNN. Finally, decision level fusion is used to increase the recognition accuracy.
	
	\begin{figure*}
		\centering
		\includegraphics[width=1.0\linewidth]{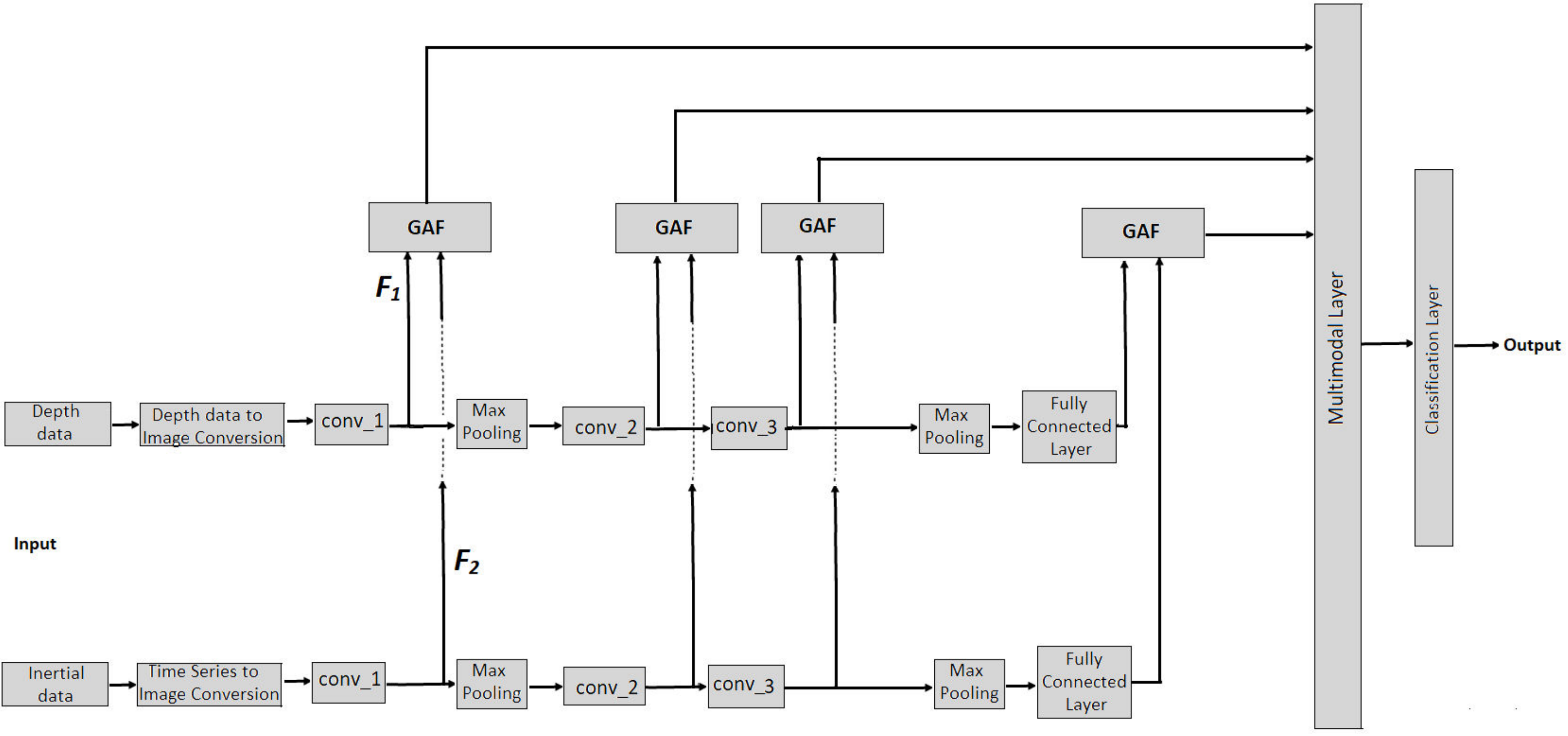}
		\caption{ Complete Overview of the Proposed Multistage Gated Average Fusion (MGAF) network.} 
		\label{fig:Proposed framework}
	\end{figure*}
	
	\begin{figure*}
		\centering
		\includegraphics[width=0.5\linewidth]{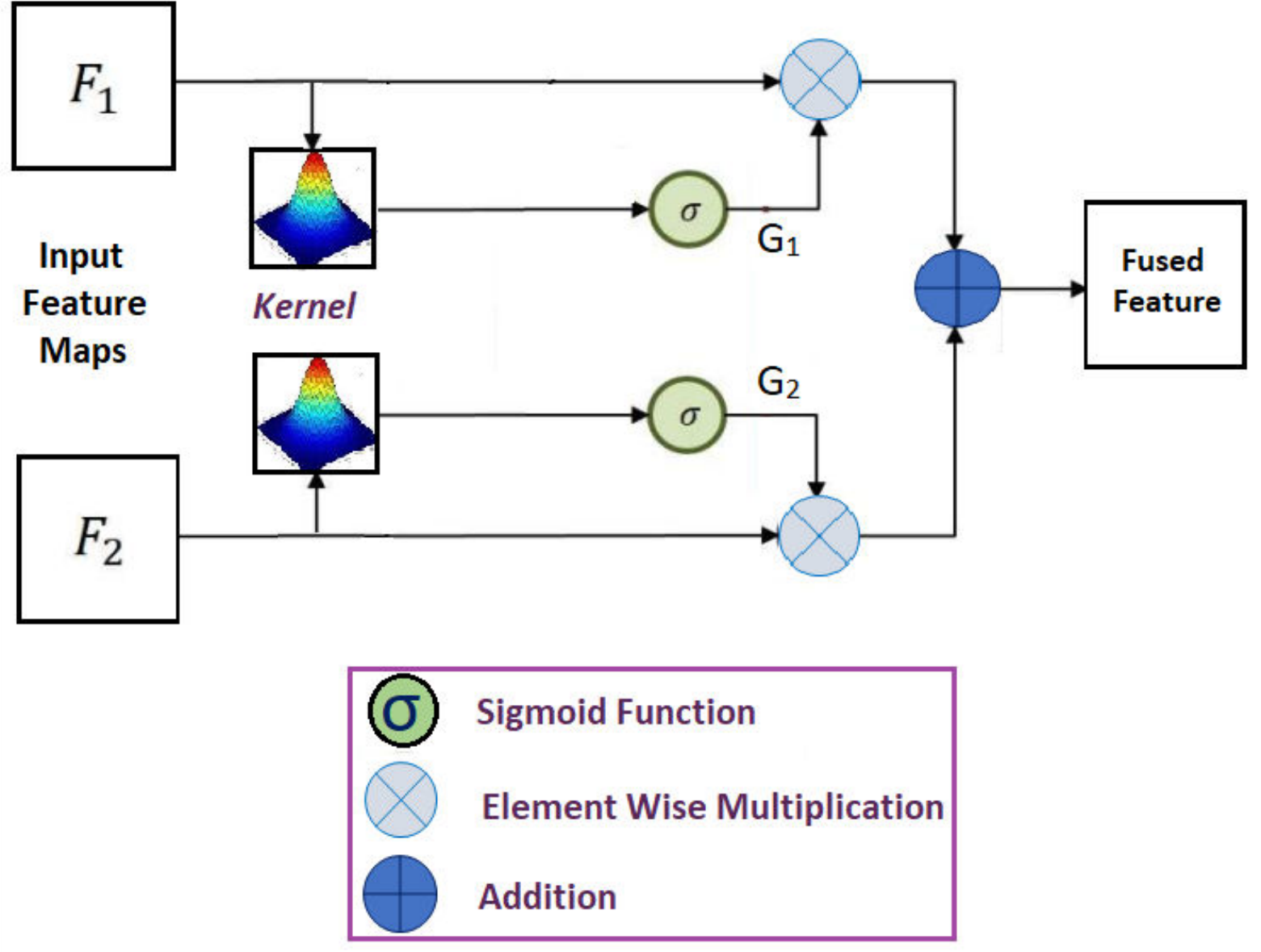}
		\caption{Architecture of the proposed Gated Average Fusion (GAF) network.}
		\label{fig:gated average fusion}
	\end{figure*}

	A common pitfall in the existing works is that the fusion strategies used are either feature level or decision level. Furthermore, these fusion tactics are exercised at a single level, thus failing to capture midlevel features which could result in better classification performance.
	
	\begin{figure*}
		\centering
		\includegraphics[width=1.0\linewidth]{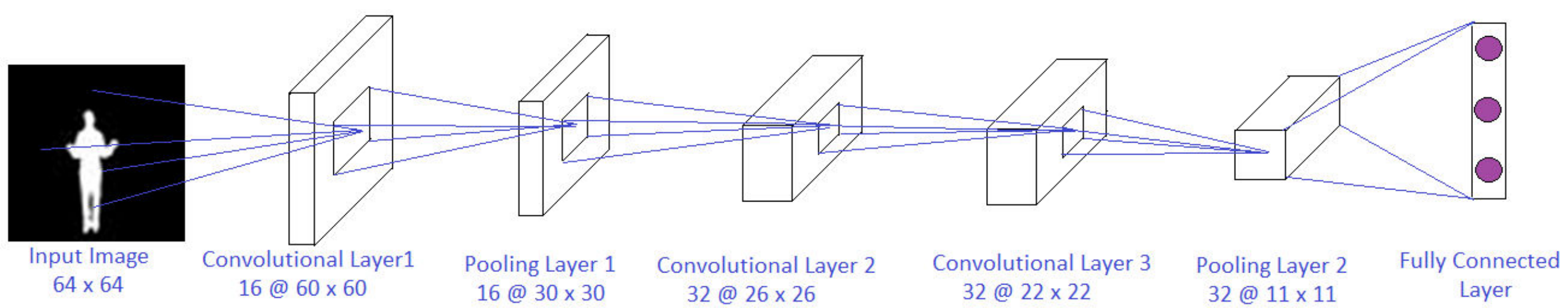}
		\caption{Architecture of CNN used in proposed MGAF.}
		\label{fig:CNN architecture}
	\end{figure*}
		
	 In~\cite{ahmad2019human}, we proposed three novel fusion frameworks that outperformed existing fusion frameworks for HAR using depth and inertial sensors. In all three fusion frameworks, fusion was performed at two stages to overcome the weaknesses of single stage fusion. Nevertheless, there are imperfections in these proposed frameworks. Firstly, the two input modalities are converted into multiple modalities, thus making computation more complex. In addition to this, features are extracted only from fully connected layers and then two stage consecutive fusion is performed, thus neglecting the complementary and unique features present in the convolutional layers of CNN to take part in fusion.
	 
Some gated fusion networks are introduced in~\cite{ren2018gated},~\cite{wu2020deepdualmapper} and~\cite{kim2018robust}. The first weakness in these fusion networks is that they apply concatenation at the early stage of fusion network, limiting the scalability and extensibility of networks for more modalities and ability to extract features from each layer of CNN. Furthermore, concatenation creates the problem of the curse of dimensionality, high computational cost, limited applicability and deterioration of accuracy~\cite{manshor2011feature}.

	 To address the shortcomings of existing works, in this paper, we propose a multistage gated average fusion (MGAF) network which has the strength and skill of extracting and fusing the features from all the layers of CNN while reducing of dimensionality as well. A significant part of the proposed MGAF network is gated average fusion (GAF) network as shown in Fig.~\ref{fig:gated average fusion}. The proposed MGAF beats or matches our previous work~\cite{ahmad2019human} with the current state-of-the-art performance on benchmark datasets, while significantly reducing the training and inference computational costs by utilizing a much smaller CNN, as described in section~\ref{sec:comparison of computational cost}.

	\section{Proposed Multistage Gated Average Fusion Framework}\label{sec:proposed method}
	
In this section, we will explain in detail the proposed multistage gated average fusion (MGAF) network shown in Fig.~\ref{fig:Proposed framework}.

 At the input of the proposed multistage gated average fusion (MGAF) network shown in Fig.~\ref{fig:Proposed framework}, both modalities are converted into images. Following our recent work in~\cite{ahmad2019human}, we convert the depth data into sequential front view images (SFI) and inertial data into signal images (SI). Two CNNs of same architecture, shown in Fig.~\ref{fig:CNN architecture}, are trained on these images. Learned features extracted from all convolutional layers and first fully connected layer are fused using gated average fusion (GAF) network shown in Fig.~\ref{fig:gated average fusion} and  explained in section~\ref{sec:GAF}.

\subsection{Formation of Sequential Front View Images}\label{sec:SFIs}

\begin{figure}[h]
	\centering
	\includegraphics[width=1.0\linewidth]{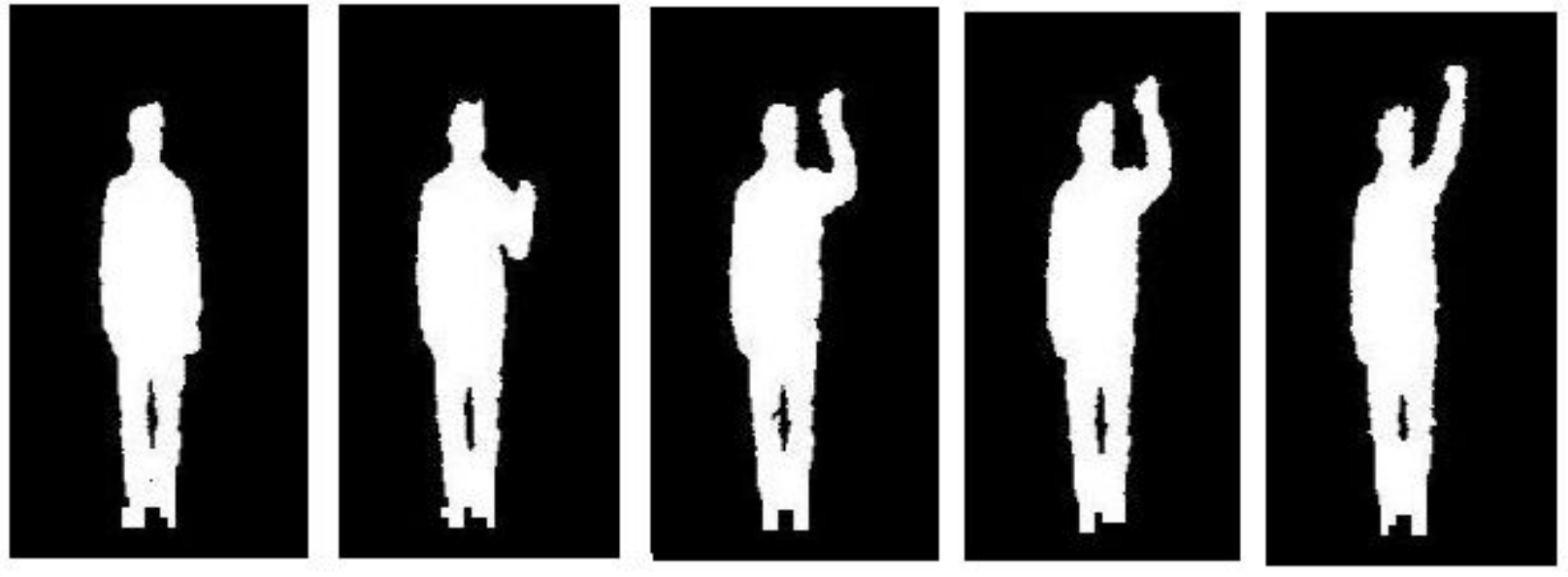}
	\caption{Five samples of Sequential Front View Images of Hand Catch}
	\label{fig:sequential front view images}
\end{figure}

At the input of the proposed MGAF, we convert depth data into images called sequential front view images (SFI) as shown in Fig.~\ref{fig:sequential front view images}. The depth modality provides 3D action information for generating front, top and side view of depth motion maps. It is experimentally proved that only front view information can be effectly utilized to recognize the actions as the fusion of front view with other views doesnot significantly increase the recognition accuracy~\cite{ahmad2019human}. For making sequential front view images (SFI), each depth sequence has the following dimension.

$Depth = rows\times  columns\times  number of frames$.

The number of SFI formed from each depth sequence is equal to the number of frames with size $rows\times columns$. We resize the image to 64 x 64 to train on CNN shown in Fig.~\ref{fig:CNN architecture}.
During fusion, supplementary information is provided by the inertial dataset. Generating SFI using only one view reduces the computational cost. SFI are like the motion energy images and motion history images introduced in~\cite{bobick2001recognition}. These SFIs provide aggregate knowledge about the dynamics of action.   

\subsection{Formation of Signal Images}\label{sec:signal images}
The combination of accelerometer and gyroscope provide multivariate time series data consists of six sequences of signals : three acceleration and three angular velocity sequences.

\begin{figure}[h]
	\centering
	\includegraphics[width=1.0\linewidth]{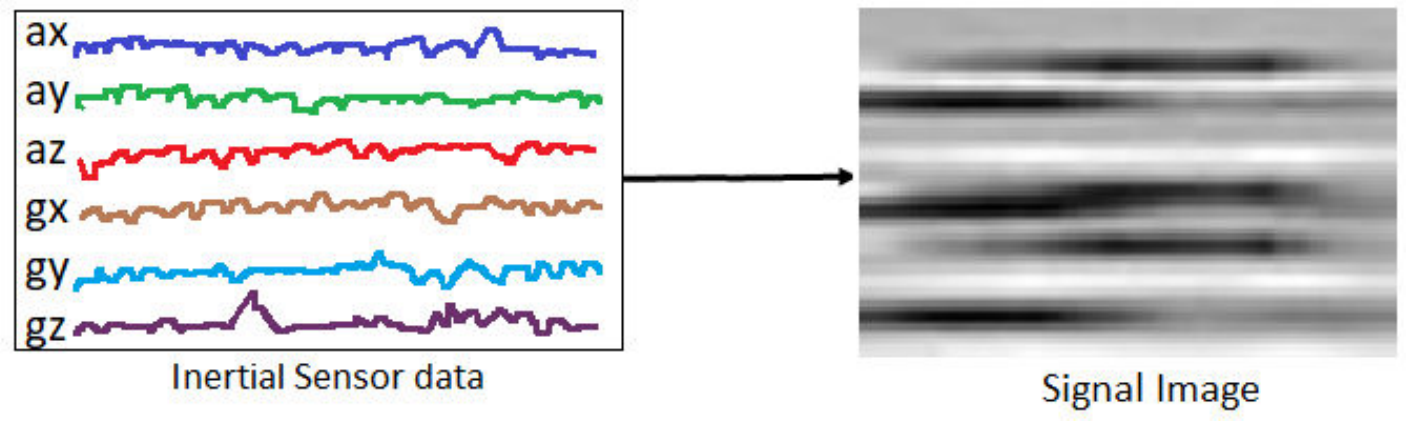}
	\caption{Formation of Signal Image from Inertial sensor data}
	\label{fig:conversion}
\end{figure}

We converted six sequences into 2D images called signal images based on the algorithm in~\cite{jiang2015human}. The conversion of time series data to signal image is shown in Fig.~\ref{fig:conversion}. Signal image is obtained through row-by-row stacking of given six signal sequences in such a way that each sequence appears alongside to every other sequence.  The signal images are formed by taking advantage of the temporal correlation among the signals. 

Row wise stacking of six sequences has the following order.

\textit{123456135246142536152616}

Where the numbers \textit{1} to \textit{6} represent the sequence numbers in a raw signal. We made sure that every sequence is a neighbor of  every other sequence only once. Using full sequence permutation order will cause redundant information in the image and that may lead to the overfitting of the CNN during training. Thus the final width of signal image becomes 24. 

\begin{figure}[h]
	\centering
	\includegraphics[width=1.0\linewidth]{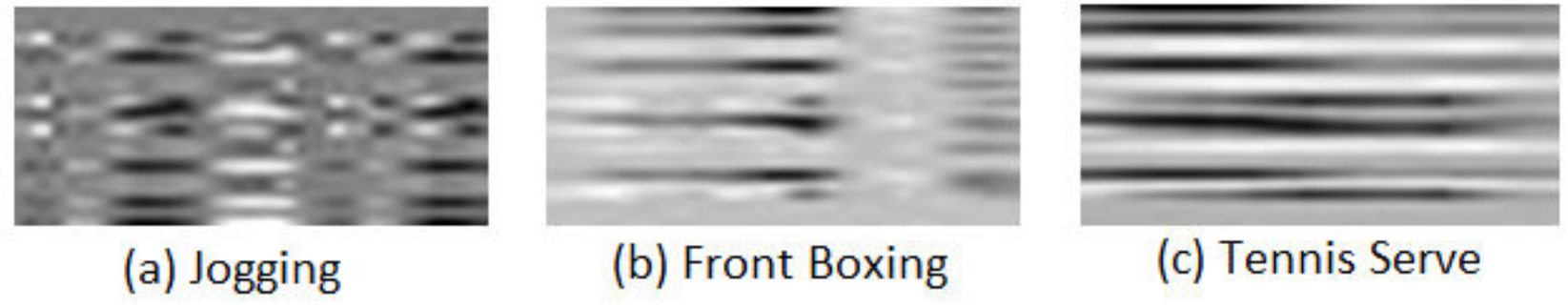}
	\caption{Signal Images of three different actions}
	\label{fig:signal images}
\end{figure}

Length of signal image is decided by making use of sampling rate of datasets which is 50Hz for our two datasets. Therefore, to capture granular motion accurately and to facilitate the design of CNN, the length of the signal image is finalized as 52, resulting in a final image size of 24 x 52. These signal images are shown in Fig.~\ref{fig:signal images} and it is observed that each signal image shows unique pattern and thus enhance the performance of classification. Finally, we resize the images to 64 x 64.
	
\subsection{Architecture of CNN}

Architecture of CNN used in proposed MGAF is shown in Fig.~\ref{fig:CNN architecture}. CNN consists of three convolutional layers, two pooling layers, and a fully connected layer. The first convolutional layer has 16 kernels of size 5x5, followed by pooling layer of size 2x2 and stride 2.
Second and third convolutional layers have 32 kernels of size 5x5 followed by 2x2 pooling layer with stride 2.

\subsection{Gated Average Fusion (GAF) Network}\label{sec:GAF}

The structure of the gated average fusion (GAF) network is shown in Fig.~\ref{fig:gated average fusion}. The input to GAF network are the feature maps extracted from the layers of CNN of each modality. The size of feature maps extracted from convolutional layer is 4-dimensional (4D) and is given by

$Size\: of feature\: map = a \times b \times n \times N$
\vspace{0.2cm}

where,

\vspace{0.1cm}

$a$ = Length of the feature map

$b$ = Width of the feature map

$n$ = Number of feature maps

$N$ = Number of training samples per batch
\vspace{0.1cm}    

To make the computation meaningful, uncomplicated and to make 2-dimensional (2D) convolution possible with high boost kernel, we transform the 4D features into 2D by following method.

\vspace{0.1cm}    

$Size\: of 2D\: feature\: map = N \times M$

\vspace{0.2cm}       

where,

\vspace{0.1cm}

$N$ = Number of training samples per batch

$M = a \times b \times n$

\vspace{0.15cm}

Let $F_1$ and $F_2$ are the feature maps of size $N \times M$ obtained from the CNNs of each modalitity respectively. These feature maps are now convolved with high boost kernel $K$. During convolution, the size of feature maps remain the same.

The kernel used for convolution with feature maps is a high boost filter that highlights the fine details of feature maps and assigns proper boosted value to each pixel of feature map according to its significance. High boost filter has been used in~\cite{alirezanejad2014effect} for watermark recovery in spatial domain by enhancing the dissimilarity between watermarked and unwatermarked parts of the image. High boost kernel can be of size 1D and 2D. However, when used as a separable filter i.e when used to enhance the important specifics of the image, it works better in 2D form. Thus, to have a meaningful and effective convolution with 2D kernel, the size of image or feature map should also be 2D. High boost filter is a second order derivative filter obtained by subtracting the low pass filtered version of the image from the scaled input image. 

\begin{equation}\label{eq:highboostequation}
f_{hb}(x,y) = Af(x,y) - f_{lp}(x,y)
\end{equation}

where $Af(x,y)$ and $f_{lp}(x,y)$ are the scaled and low pass versions of original image $f(x,y)$

The general form of high boost filter is given by

\vspace{0.1cm}

\begin{equation}
K=
\begin{bmatrix}\label{eq:highboostmatrix}

-1 & -1 & -1 \\
-1 & A+8 & -1 \\
-1 & -1 & -1

\end{bmatrix}
\end{equation}	

 where $A$ is called amplification factor that controls the amount of weight given to the feature maps during convolution.

 For $A$ = 1, we observe optimal filter performance. Higher values of $A$ causes less boosting. 
 
Thus, we empirically select the following high boost kernel that emphasizes the important specifics. 

\vspace{0.1cm}

\begin{equation}\label{eq:final highboost eqn}
K=
\begin{bmatrix}

-1 & -1 & -1 \\
-1 & 9 & -1 \\
-1 & -1 & -1

\end{bmatrix}
\end{equation}	
 
 High boost filter enhances high frequency components while still preserving the low frequency components. 

After convolution with kernel, sigmoid function is applied to generate the gated values $G_1$ and $G_2$ respectively as shown in Fig.~\ref{fig:gated average fusion}. Sigmoid function squishes the values between 0 and 1.
Finally, the gated values $G_1$ and $G_2$ are element-wise multiplied with feature maps $F_1$ and $F_2$ respectively to perform gated avearge fusion and to generate fused feature map. This gated fusion is performed with all feature maps obtained from three convolutional layers and first fully connected layer of the CNNs of both modalities as shown in Fig.~\ref{fig:Proposed framework}. The operation of GAF network is summarized by the following mathematical equations. 

\vspace{0.1cm}    

\begin{equation} \label{eq:first equation}
	G_1 = \sigma(F_1 \circledast\ K)
\end{equation}

\begin{equation} \label{eq:second equation}
	G_2 = \sigma(F_2 \circledast\ K)
\end{equation}

\begin{equation} \label{eq:third equation}
	F_f(j)  = G_1 \odot F_1(j) + G_2 \odot F_2(j)
\end{equation}

\vspace{0.2cm}       

where,
\vspace{0.1cm}

$\sigma(z) \triangleq \frac{1}{1+e^{-z}}$ : sigmoid function.

\vspace{0.1cm}

$y \circledast z$ : convolution operation

\vspace{0.1cm}

$y \odot z$ : element-wise multiplication

\vspace{0.1cm}

$F_i(j)$ : $j$th feature map of $i$th modality 

\vspace{0.1cm}

$F_f(j)$ : $jth$ fused feature map

\vspace{0.2cm}

The working of equations~\ref{eq:first equation} and~\ref{eq:second equation} can be thought of as an input gate of LSTM that decides which values will be updated by transforming the values to be between 0 and 1. 0 means not important, and 1 means most important.

The GAF network scales very well for multimodal fusion, since the gated average fusion output size remains the same regardless of the number of modalities. If $F_1,F_2,F_3,...,F_N$ are feature maps of $N$ input modalities each of size $N\times M$, then the fused feature map $F_f$ will also be of size $N\times M$. This can also be understood by transforming equation~\ref{eq:third equation} for $N$ modalities.

\begin{equation} \label{eq:fourth equation}
	F_f  = G_1 \odot F_1 + G_2 \odot F_2 + G_3 \odot F_3 + .... + G_N \odot F_N
\end{equation}

\subsection{MGAF-Architecture}

 The fused features generated by the gated average fusion (GAF) network from different CNN layers are collected at multimodal layer and concatenated and finally sent to the SVM classifier for classification task. The multimodal layer serves as input to the SVM classifier. The SVM classifier is trained separately from the CNNs. CNNs are used as  feature extractors, SVM is employed for the final classification. We used support vector machine (SVM) as a classifier since we experimentally proved in our previous work~\cite{ahmad2018towards} that SVM performs better than softmax, which is typically built into any CNN framework.
 Softmax classifier reduces the cross entropy function while SVM employs a margin based function. The more rigorous nature of classification is the reason of better performance of SVM over softmax.
  
Although we explain MGAF with two modalities, the proposed MGAF can be conveniently out-stretched to more than two modalities without increasing the size of the fused feature map. This is due to the fact that GAF network of MGAF architecture generates fused features by performing gated average fusion which keeps the size of fused feature map equal to the size of single input feature map and thus avoid the curse of dimensionality which is commonly observed in case of concatenation fusion. To show extensibility, we also perform experiments using three modalities and display results in section~\ref{sec:Experiment and results}.

\begin{table}[h]
	\caption{Dataset Information} 	
	\begin{adjustbox}{width=\columnwidth,center}
		\renewcommand\arraystretch{1.3}
		
		\centering
		
		\begin{tabular}{c c c c c}
			\hline\hline
			\textbf{Dataset}     & \textbf{Sampling Rate}    &\textbf{ Modality}    & \textbf{Training Samples}   & \textbf{Test Samples}   \\
			\hline\hline		
			\multirow{3}*{UTD MHAD}
			& \multirow{2}*{50}
			& Depth    & 46639    & 11660    \\
			\cline{3-5}
			&           & \multirow{1}*{Inertial}
			& 11031    & 2745    \\\hline
			
			\multirow{3}*{Berkeley MHAD}
			& \multirow{2}*{30}
			& Depth    & 26400    & 6500    \\
			\cline{3-5}
			&           & \multirow{1}*{Inertial}
			& 2612   & 653   \\\hline
			
			\multirow{3}*{UTD Kinect V2}
			& \multirow{2}*{50}
			& Depth    & 14098    & 3524   \\
			\cline{3-5}
			&           & \multirow{1}*{Inertial}
			& 3532  & 884   \\ 
			\hline\hline
			
		\end{tabular}
	\end{adjustbox}
	\label{dataset information}
\end{table}

\section{Experiments and Results} \label{sec:Experiment and results}
We experiment on three publicly available multimodal human action datasets, namely, UTD-MHAD~\cite{chen2015utd}, Berkeley MHAD~\cite{ofli2013berkeley} and UTD Kinect-V2 dataset~\cite{chen2016fusion}. We used subject specific setting for experiments on all datasets. In subject specific setting, training and testing sets are split randomly across all subjects.

For all experiments, we split the datasets into training and testing samples by randomly splitting 80\% data into training and 20\% data into testing samples. We ran the random split 20 times and report the average accuracy. The number of training and testing samples after splitting are shown in Table~\ref{dataset information}. 

\subsection{Ablation Study}\label{sec:ablation study}

 In order to show the significance of the proposed MGAF, we carry out ablation study on all datasets using concatenation, average fusion and gated fusion without kernel. Average fusion is performed by assigning the value one to both gated outputs i.e $G_1=1$ and $G_2=1$ in the GAF network.
 In gated fusion without kernel, the sigmoid function is applied directly to input feature maps $F_1$ and $F_2$ and the gated values are calculated by $G_1=\sigma(F_1)$ and $G_2=\sigma(F_2)$. The results of ablation study are reported in Tables~\ref{Tab:experiments on utdmahad},~\ref{Tab:experiments on berkeley mhad} and~\ref{Tab:experiments on kinect-v2} along with proposed MGAF. The results of ablation study are explained in detail in section~\ref{sec:discussion on results}.
 
Two Convolutional neural networks of same architecture shown in Fig.~\ref{fig:CNN architecture}, are trained on SFI and SI as shown in Fig.~\ref{fig:Proposed framework}. In addition to this, momentum of 0.9, initial learn rate of 0.005, $L_2$ regularization of 0.004 and mini-batch size of 64 is used to control overfitting. We reached these value through using the grid search method. We conduct our experiments on Matlab R2018b on a desktop computer with NVIDIA GTX-1070 GPU. We train CNNs on images till the validation loss stops decreasing further.

\begin{table*}
	\vspace{0.3cm}
	\caption{Experimental results on UTD MHAD}
	\begin{adjustbox}{width=\columnwidth,center}
		\renewcommand\arraystretch{1}
		\centering
		\begin{tabular}{c c c c c}	
			\hline\hline
			\textbf{Modalities} & \textbf{Concatenation} & \textbf{\makecell{Average \\ Fusion}} & \textbf{\makecell{Gated Fusion\\ without Kernel}}&\textbf{\makecell{Proposed \\ Gated Average Fusion}} \\\hline\hline
			SFI + SI  &  97.8 & 99.05 &99.13 & 99.3 \\\hline
			SFI + SI + SFI (Prewitt)  & 94.7 & 99.1 & 99.22 & 99.4 \\\hline
			SI(Spatial) + SI(DFT) + SI (GWT) & 87.2 & 95 & 96.4 &96.8\\\hline\hline						
		\end{tabular}
	\end{adjustbox}
	
	\label{Tab:experiments on utdmahad}
\end{table*} 

\begin{table*}[h]
	\caption{Experimental results on Berkeley MHAD}
	\begin{adjustbox}{width=\columnwidth,center}
		\renewcommand\arraystretch{1}
		\centering
		\begin{tabular}{c c c c c}	
			\hline\hline
			\textbf{Modalities} & \textbf{Concatenation} & \textbf{\makecell{Average \\ Fusion}} & \textbf{\makecell{Gated Fusion\\ without Kernel}}&\textbf{\makecell{Proposed \\ Gated Average Fusion}} \\\hline\hline
			SFI + SI  &  99.2 & 99.4 & 99.5 & 99.85 \\\hline
			SFI + SI + SFI (Prewitt)  & 99 & 99.6 &99.7& 99.9 \\\hline
			SI(Spatial) + SI(DFT) + SI (GWT) & 91.2 & 98.2 &98.6& 98.9\\\hline\hline						
		\end{tabular}
	\end{adjustbox}
	
	\label{Tab:experiments on berkeley mhad}
\end{table*} 

\begin{table*}[h]
	\caption{Experimental results on Kinect-V2 dataset}
	\begin{adjustbox}{width=\columnwidth,center}
		\renewcommand\arraystretch{1}
		\centering
		\begin{tabular}{c c c c c}	
			\hline\hline
			\textbf{Modalities} & \textbf{Concatenation} & \textbf{\makecell{Average \\ Fusion}} & \textbf{\makecell{Gated Fusion\\ without Kernel}}&\textbf{\makecell{Proposed \\ Gated Average Fusion}} \\\hline\hline
			SFI + SI  &  97 & 99.3 &99.44& 99.8 \\\hline
			SFI + SI + SFI (Prewitt)  & 96.2 & 99.7 &99.74 & 99.9 \\\hline
			SI(Spatial) + SI(DFT) + SI (GWT) & 91 & 98.8 & 99.1&99.4\\\hline\hline						
		\end{tabular}
	\end{adjustbox}
	\vspace{0.7cm}
	\label{Tab:experiments on kinect-v2}
\end{table*}

\begin{table*}[h]

	\begin{adjustbox}{width=\columnwidth,center}
		\begin{tabular}{c c c c}
			
			\hline\hline
			\textbf{Fusion Frameworks} & \textbf{Training Speed ($minutes$)} & \textbf{Inference Speed ($\mu$s)} & \textbf{Training Parameters}  \\\hline\hline 
			Deep Multistage Feature Fusion Framework~\cite{ahmad2019human} & 268 & 158 & 49232121  \\\hline	
			Deep Hybrid Fusion Framework~\cite{ahmad2019human} & 268 & 235 & 49232121  \\\hline
			Computationally Efficient Fusion Framework~\cite{ahmad2019human} & 145 & 147& 25180266\\\hline
			\textbf{Proposed MGAF (SFI + SI)}  & \textbf{58} & \textbf{132} & \textbf{1229814}\\\hline\hline
			
		\end{tabular}
	\end{adjustbox}
	\caption{Comparison of Computational Cost of Proposed MGAF with previous state of art on UTD-MHAD Dataset}
	\vspace{0.3cm}
	\label{tab:computationalcomparison}
\end{table*}

\subsection{UTD-MHAD Dataset}\label{sec:UTDMHAD dataset}
The UTD-MHAD dataset consists of both depth and inertial components and is collected in an indoor environment by 12 subjects with each subject repeating the actions 4 times.  The
dataset contains 27 actions. 

The inertial component of UTD-MHAD dataset is very challenging for deep network training due to few flaws. The first drawback is that inertial sensor was worn either on volunteer's right wrist or right thigh depending upon the nature of action. Hence the  sensor is worn only on two positions for collecting data of 27 actions which is insufficient to capture all the dependencies and characteristics of data. Furthermore, the number of data samples after converting to SI are only 2206, which is not enough to train a deep network properly. 

To combat these challenges, we perform data augmentation on signal images to increase the number of samples by applying the data augmentation techniques discussed in~\cite{ahmad2018towards}. The number of samples shown in Table~\ref{dataset information} are obtained after augmentation of signal images.

We perform experiments on our proposed MGAF using UTD-MHAD datsets with same number of samples shown in Table~\ref{dataset information} and with same parameters of CNN described in section~\ref{sec:ablation study}. The results of all experiments in terms of recognition accuracies are presented in Table~\ref{Tab:experiments on utdmahad}. These results are discussed in detail in sections~\ref{sec:discussion on results} and~\ref{sec:feature visualization}.

Since proposed MGAF is extendable to more than two modalities as explained in section~\ref{sec:proposed method}, we also perform experiments by creating more modalities to validate the usability of the proposed MGAF.  We create another modality called SFI (Prewitt) from depth data by convolving SFI with Prewitt filter as described in our previous work in~\cite{ahmad2019human}. The results obtained by experimenting with three modalities i.e SFI, SI and SFI (Prewitt) are also reported in Table~\ref{Tab:experiments on utdmahad}.

We also made inertial data multimodal by creating two more modalities using Discrete Fourier transform (DFT) and Gabor Wavelet transform (GWT) as illustrated in~\cite{ahmad2019multidomain}. We obtained recognition accuracy of 96.8\% as shown in the last column of Table~\ref{Tab:experiments on utdmahad} beating previous state of art 95.8\% as reported in~\cite{ahmad2019multidomain}.

The comparison of recognition accuracies by using only two modalities i.e depth and inertial with previous state-of-the-art on UTD-MHAD is shown in Table~\ref{tab:UTDMHAD comparisonTabe}.

Most of the work in Table~\ref{tab:UTDMHAD comparisonTabe} used subject specific setting except in~\cite{bulbul2015dmms},~\cite{dawar2019data} and~\cite{dawar2018action}. We used subject specific setting since this is a most commonly used~\cite{chen2016fusion}. We compare the results on the basis of modalities i.e depth and inertial.

\begin{table}[h]
	\centering
		\begin{tabular}{c c}
			
			\hline\hline 
			\textbf{Previous Methods} & \textbf{Accuracy\%}  \\ \hline\hline 
			C.chen et al.~\cite{chen2015utd}       &      97.1 \\\hline
			Bulbul et al.~\cite{bulbul2015dmms}    &      88.4 \\\hline
			N.Dawar et al. ~\cite{dawar2019data}             &      89.2 \\\hline
			N.Dawar et al. ~\cite{dawar2018action}            &      92.8 \\\hline
			Chen et al.~\cite{chen2016real}        &      97.2 \\\hline
			Ehatisham et al.~\cite{ehatisham2019robust} &   98.3\\\hline
			Z.Ahmad et al. ~\cite{ahmad2018towards}              &      98.4 \\\hline
			Mahjoub et al.~\cite{mahjoub2018efficient}        &      98.5 \\\hline
			Z.Ahmad et al.~\cite{ahmad2019human}              &    99.2 \\\hline
		    \textbf{Proposed MGAF (SFI + SI)}                            &   \textbf{99.3} \\
			
			\hline\hline
			
		\end{tabular}
	\caption{ Comparison of Accuracies of the proposed MGAF network with previous methods on UTD-MHAD dataset using depth and inertial modalities.}
	\label{tab:UTDMHAD comparisonTabe}
\end{table}

\subsection{Berkeley Multimodal Human Action Dataset}\label{sec:Berkeley MHAD}
The Berkeley MHAD is another multimodal dataset containing both depth and inertial modalities. The dataset contains 11 actions performed five times by seven male and five female subjects.  

 Inertial part of the dataset contains six accelerometers and each generates three sequences. For generating signal images explained in section~\ref{sec:signal images}, we need six sequences in a row. Thus we used two accelerometer ($A_1$ and $A_4$) and stacked them row wise to make six sequences. The reason for selecting $A_1$ and $A_4$ is that they are worn on the left wrist and right hip, respectively and are able to generate more useful information than those worn on both ankles~\cite{chen2015improving}.
We performed experiments on Berkeley MHAD dataset using proposed MGAF network with same setting shown in Table~\ref{dataset information}, with the same parameters mentioned in section~\ref{sec:ablation study} and using the same modalities as that for UTD-MHAD in section~\ref{sec:UTDMHAD dataset}.
The results of experiments on the dataset and their comparison with previous state-of-the-art are shown in Tables~\ref{Tab:experiments on berkeley mhad} and~\ref{tab:comparison on Berkeley MHAD dataset} respectively. Detailed discussion on experimental results is presented in sections~\ref{sec:discussion on results} and~\ref{sec:feature visualization}.

\begin{table}[h]
		\begin{tabular}{c c}
			
			\hline\hline 
			\textbf{Previous Methods} & \textbf{Accuracy\%}  \\\hline\hline
			F. Ofli et al.~\cite{ofli2013berkeley}       &      97.81 \\\hline
			Alireza Shafaei et al.~\cite{shafaei2016real} &     98.1   \\\hline
			Earnest Paul Ijjina et al.~\cite{ijjina2014human} &  98.38 \\\hline
			Chen Chen et al.~\cite{chen2015improving}       &      99.54 \\\hline
			Z.ahmad et al.~\cite{ahmad2019human}            &      99.8  \\\hline
			 \textbf{Proposed MGAF (SFI + SI)}                            &   \textbf{99.85} \\
			\hline\hline
		\end{tabular}
	\caption{Comparison of Accuracies of proposed MGAF network with previous methods on Berkeley MHAD dataset using depth and inertial modalities.}
	\label{tab:comparison on Berkeley MHAD dataset}
\end{table}

\begin{figure*}
		
	\centering
	\includegraphics[width=1.0\linewidth]{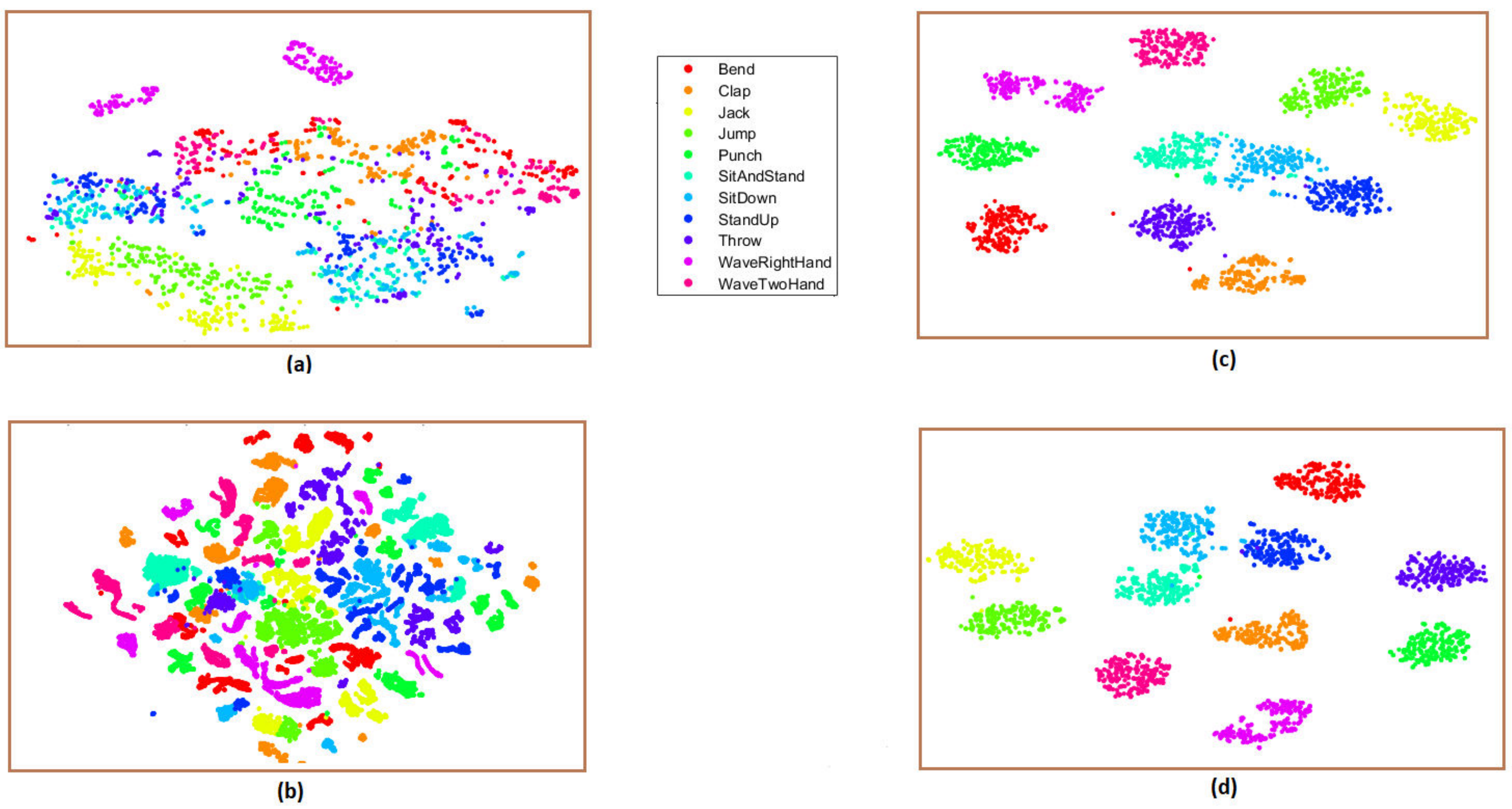}
	\caption{Features Visualization of Berkeley MHAD. (a) Features without fusion. (b) Features grouping after concatenation. (c) Features grouping after average fusion. (d) Features grouping after gated average fusion.}
	\label{fig:feature visualization}
\end{figure*}

\subsection{UTD kinect-V2 Dataset}\label{sec:Kinect2 dataset}

Kinect-V2 action dataset is another publicly available dataset that contains both depth and inertial data. It is a new dataset using the second generation of kinect. It contains 10 actions performed by six subjects with each subject repeating the action 5 times. 

We performed experiments on Kinect V-2 dataset with same settings shown in Table~\ref{dataset information}, with the same parameters of CNN described in section~\ref{sec:ablation study} and using the same modalities as that for UTD-MHAD in section~\ref{sec:UTDMHAD dataset} and Berkeley MHAD in section~\ref{sec:Berkeley MHAD}.
The results in terms of recognition accuracies for the dataset and their comparison with previous state-of-the-art are shown in Tables~\ref{Tab:experiments on kinect-v2} and~\ref{tab: comparison on Kinect 2 dataset} respectively. Detailed analysis on experimental results is provided in section~\ref{sec:discussion on results}.

\begin{table}[h]
		\begin{tabular}{c c}
			\hline\hline 
			\textbf{Previous Methods} & \textbf{Accuracy\%}  \\\hline\hline
			Chen et al.~\cite{chen2016fusion}       &      99.5 \\\hline
			Z.Ahmad et al.~\cite{ahmad2018towards}  &      99.8  \\\hline
			Z.Ahmad et al.~\cite{ahmad2019human}  &       99.8  \\\hline
			 \textbf{Proposed MGAF (SFI + SI)}               &   \textbf{99.8} \\
			\hline\hline
			
		\end{tabular}
	\caption{Comparison of Accuracies of proposed MGAF network with previous methods on Kinect V2 dataset using depth and inertial modalities.}
	\vspace{0.3cm}
	\label{tab: comparison on Kinect 2 dataset}
\end{table}

\let\thefootnote\relax\footnote{Gated average fusion code can be found at https://github.com/zaamad/Gated-Average-Fusion}

\subsection{Discussion on Results}\label{sec:discussion on results}

Experimental results show that the concatenation performance is poor compared to other two methods, average fusion and gated average fusion, and it gets worse with the increase in modalities as shown in Tables~\ref{Tab:experiments on utdmahad},~\ref{Tab:experiments on berkeley mhad},~\ref{Tab:experiments on kinect-v2}. Concatenation of features from each convolutional layer with different visual properties and dynamic ranges results in very high dimensionality of feature space  that contains inconsequential features and produces repeated and false information that leads to the curse of dimensionality and in turns additional computational complexity and degradation of classification accuracy~\cite{akbas2007automatic}.
 Moreover, concatenation works better when features from different modalities are strongly correlated~\cite{cai2014multi}.

Tables~\ref{tab:computationalcomparison},~\ref{tab:UTDMHAD comparisonTabe},~\ref{tab:comparison on Berkeley MHAD dataset} and~\ref{tab: comparison on Kinect 2 dataset} justify the superiority of the proposed method over the previous state-of-the-art. An important point to mention is that we achieve higher accuracy using less computational resources as shown by the inference speeds in Table~\ref{tab:computationalcomparison}. Action recognition is likely to be a real-time application, so we believe lower inference speed while surpassing the state-of-the-art accuracy is a strong contribution. The theoretical framework behind MGAF also has the potential to be applied in other application areas, since MGAF can separate the feature space better, as shown by the feature visualization in Fig.~\ref{fig:feature visualization}.

In the proposed method, we extract features from all convolutional layers and first fully connected layer of the CNN. During multimodal learning, the correlation among the features increases as we move from lower layers of CNN to higher.  We experimentally prove this fact by calculating the normalized correlation coefficients for each dataset as shown in Table~\ref{tab:NCC}.

\begin{table}[h]
	
	\centering
	\scalebox{0.9}{
		\begin{tabular}{c c c c}	
			\hline\hline 
			\textbf{Datasets} & \textbf{conv1} & \textbf{conv2} & \textbf{conv3} \\\hline\hline 
			UTD-MHAD  & 0.00032 & 0.0026 & 0.033 \\\hline
			Berkeley MHAD  & 0.0037 & 0.012 & 0.015 \\\hline
			Kinect V2 & 0.0003 & 0.0016 & 0.0089\\\hline\hline						
	\end{tabular}}
	\caption{Normalized cross correlation coefficients between features extracted from all three convolutional layers of CNN.}
	\label{tab:NCC}
\end{table} 	

The normalized correlation coefficients are calculated using the following equation.

\begin{equation}
NCC(F_1,F_2)= \frac{1}{P}\frac{\sum_{x,y}(F_1-\bar{f_1})(F_2-\bar{f_2})}{\sigma_1 \sigma_2}
\end{equation}

where,

$\bar{f_1}$ and $\bar{f_2}$ are the average pixel intensities of each feature map.
$\sigma_1$ and $\sigma_2$ are the standard deviation of pixel intensities of each feature map. $P$ is the total number of pixels in the feature map. The NCC value is between +1 and -1. If two feature maps are highly correlated, NCC=1, and NCC = 0 if they are completely uncorrelated.

In Table~\ref{tab:NCC}, conv1, conv2 and conv3 denote the first, second and third convolutional layers of CNN shown in Fig.~\ref{fig:CNN architecture}.

As we can see since higher layers have higher NCC values, concatenation performs better if the  features are extracted from the higher layers of CNN~\cite{ahmad2018towards}.

In order to alleviate these limitations, we presented in this paper a novel fusion framework that selected the most consequential and meaningful features while reducing dimensionality as well. As a result, the proposed method outperforms the state-of-the-art methods on all three benchmark datasets, as can be seen from Tables~\ref{tab:UTDMHAD comparisonTabe},~\ref{tab:comparison on Berkeley MHAD dataset} and~\ref{tab: comparison on Kinect 2 dataset}. While the performance of the proposed MGAF is very close to our recently proposed frameworks in~\cite{ahmad2019human}, as we show in Section~\ref{sec:comparison of computational cost}, MGAF achieves state-of-the-art performance with a significantly reduced training and inference time due to the gating mechanism and dimensionality reduction.

\subsection{Qualitative Analysis of Results}\label{sec:feature visualization}

For qualitative analysis, we construct feature visualizations for Berkeley MHAD by reducing high dimensional features to 2 dimensional using t-SNE~\cite{maaten2008visualizing} as shown in Fig.~\ref{fig:feature visualization}. In~Fig.~\ref{fig:feature visualization}(a), features are shown without fusion and we observe that the classes seem to be inseparable. Fig.~\ref{fig:feature visualization}(b), (c) and (d) are feature visualizations after concatenation, average fusion and gated average fusion respectively. In concatenation, we inspect that between-class distance is still small and classes make clusters at more than one places and hence lead to misclassification.

For Average fusion, we recognize that between-classs distance has increased and the clusters are more separated. However, there are actions like \textit{sit and stand, sit down and stand up}, creating complications due to resemblance between them and thus their between-class distance is still insufficient for segregation. This problem is solved by our proposed GAF as shown in Fig.~\ref{fig:feature visualization}(d). We can now distinguish all classes clearly and clusters are more separable than before.

In Fig.~\ref{fig:feature visualization}, we can see that features are distinctly separated and these distinct features improve the performance of a classifier and thus highest recognition accuracy is achieved as compared to the previous works.
 This shows the supremacy of the proposed method. Since MGAF helps in separating the feature space in a more distinctive manner, we believe that the application of the proposed framework may help improving performance in other application areas as well.

\subsection{Comparison of Computational Cost of Proposed Method with Previous State of Art}\label{sec:comparison of computational cost}

Table~\ref{tab:computationalcomparison} shows that the proposed fusion network is computationally efficient than previous state of art in terms of training speed, inference speed and number of training parameters. In~\cite{ahmad2019human}, computational cost, especially the training speed and number of training parameters, became high due to the creation of extra modalities and use of deeper CNN. Please note that the Tables~\ref{Tab:experiments on utdmahad},~\ref{Tab:experiments on berkeley mhad} and~\ref{Tab:experiments on kinect-v2} show that by utilizing more than two modalities, the propsed MGAF can achieve an even better accuracy. However, for a balanced comparison, we only report the results from two modalities (SFI + SI) in Tables~\ref{tab:computationalcomparison},~\ref{tab:UTDMHAD comparisonTabe},~\ref{tab:comparison on Berkeley MHAD dataset} and~\ref{tab: comparison on Kinect 2 dataset}.

 Training speed is expressed in minutes and is defined as the total time to train the whole framework. Inference speed is expressed in microseconds ($\mu$s) and is defined as a time taken by classifier to classify single test sample.

Hence Tables~\ref{tab:computationalcomparison},~\ref{tab:UTDMHAD comparisonTabe},~\ref{tab:comparison on Berkeley MHAD dataset} and~\ref{tab: comparison on Kinect 2 dataset} show that the proposed method beats the previous state-of-the-art in terms of recognition accuracies and computational cost.

\section{Conclusion}

In this paper, we proposed a novel fusion network called multistage gated average fusion (MGAF) which is capable of extracting features from all layers of CNN and fuse them using our proposed gated average fusion (GAF) network. At the input of the proposed MGAF, we transform both input modalities into images. Features are extracted from both input modalities by utilizing two CNNs of the same architecture. Extracted features are fused effectively by the GAF network which assigns gated values to input feature maps. Experimental results on three publicly available multimodal HAR datasets show the superiority of the proposed MGAF over recently proposed multimodal fusion methods for depth-inertial HAR in terms of recognition accuracy and computational cost. In our furture work we are planning to implement the proposed method on other modalities than inertial and depth. The proposed method has some limitations. If the input image size is too large or if the CNN is deeper, extracting and fusing features from all layers of the CNN will not be computationally feasible. In future, we will investigate how appropriate layers to fuse can be selected automatically in an intelligent manner.

 \bibliographystyle{IEEEtran}

  \begin{IEEEbiography}[{\includegraphics[width=0.8in,height=1in,clip]{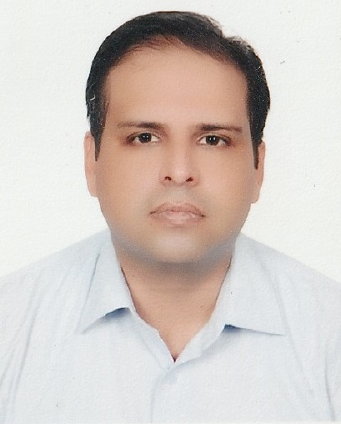}}]{Zeeshan Ahmad} received B.Eng. degree in Electrical Engineering from NED University of Engineering and Technology Karachi, Pakistan in 2001, M.Sc. degree in Electrical Engineering from National University of Sciences and Technology Pakistan in 2005 and MEng. degree in Electrical and Computer Engineering from Ryerson University, Toronto, Canada in 2017.  He is currently pursuing Ph.D. degree with the Department of Electrical and Computer Engineering, Ryerson University, Toronto, Canada. His research interests include Machine learning, Computer vision, Multimodal fusion, signal and image processing.
  \end{IEEEbiography}

\begin{IEEEbiography}[{\includegraphics[width=0.8in,height=1in,clip]{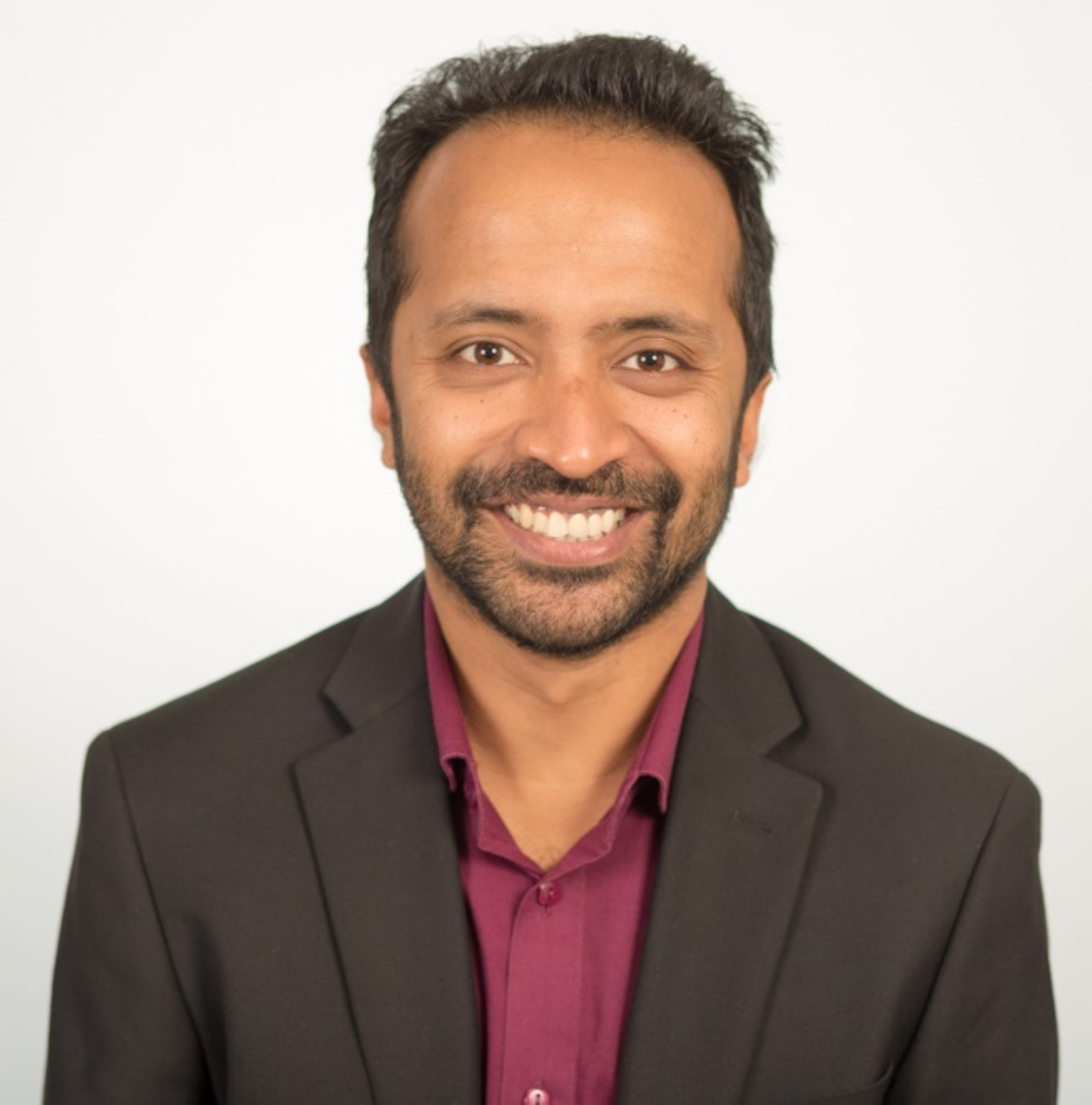}}]{Naimul Khan} is an assistant professor of Electrical and Computer Engineering at Ryerson University, where he co-directs the Ryerson Multimedia Research Laboratory (RML). His research focuses on creating user-centric intelligent systems through the combination of novel machine learning and human-computer interaction mechanisms.  He is a recipient of the best paper award at the IEEE International Symposium on Multimedia, the OCE TalentEdge Postdoctoral Fellowship, and the Ontario Graduate Scholarship. He is a senior member of IEEE and a member of ACM.
\end{IEEEbiography}
  
\end{document}